\documentclass[letterpaper, 10 pt, conference]{ieeeconf}  

\IEEEoverridecommandlockouts                              
\usepackage{graphics} 
\usepackage{amssymb}  
\usepackage{textglos}
\usepackage[utf8]{inputenc} 
\usepackage[T1]{fontenc}    
\usepackage{hyperref}       
\usepackage{url}            
\PassOptionsToPackage{hyphens}{url}\usepackage{hyperref}
\usepackage[hyphenbreaks]{breakurl}
\usepackage{booktabs}       
\usepackage{amsfonts}       
\usepackage{nicefrac}       
\usepackage{microtype}      
\usepackage{lipsum}
\usepackage{tabularx} 
\usepackage{amsmath}  
\usepackage{graphicx,dblfloatfix} 
\usepackage{cite} 
\usepackage{titlesec}
\usepackage{mathtools, amssymb, nccmath}
\usepackage{wasysym}
\usepackage{algorithm}
\usepackage{algorithmic}
\usepackage{multicol}
\let\labelindent\undefined
\usepackage{enumitem}
\usepackage{float}
\usepackage{mathtools}
\usepackage{amsmath,amsfonts,amssymb}
\usepackage{cmll}
\usepackage{tensor}
\usepackage{booktabs}
\usepackage{tabularx}
\usepackage{makecell}

\usepackage{tablefootnote}
\usepackage{threeparttable}
\usepackage{multirow}

\allowdisplaybreaks[1]

\usepackage{setspace}
\usepackage[font=small,skip=5pt]{caption}

\usepackage[dvipsnames]{xcolor}

\newcolumntype{Z}{ >{\centering\arraybackslash}X }

\newcommand{\crff}{\,\overline{\!\times\!}{}^{\,*}}

\newcommand{\M}{\mathcal{M}}

\newcommand{\I}{\mathcal{I}}

\newcommand{\T}{^\top}
\newcommand{\Rten}{^{\widetilde{\mathrm{R}}}} 

\newcommand{\w}{\mathbf{w}}

\newcommand{\R}{\mathbb{R}}
\newcommand{\C}{\vC}

\newcommand{\rmvec}[1]{\boldsymbol{#1}}
\newcommand{\greekvec}[1]{\boldsymbol{#1}}

\newcommand{\B}{\boldsymbol{B}{}}
\renewcommand{\C}{\boldsymbol{C}}
\renewcommand{\M}{\boldsymbol{M}}

\renewcommand{\I}{\boldsymbol{I}{}}

\newcommand{\g}{\rmvec{g}}
\newcommand{\q}{\rmvec{q}}
\renewcommand{\b}{\rmvec{b}}
\renewcommand{\u}{\rmvec{u}}

\usepackage{framed}

\newcommand{\f}{\rmvec{f}}

\renewcommand{\v}{\rmvec{v}}

\renewcommand{\a}{\rmvec{a}}

\newcommand{\qd}{\dot{\q}}
\newcommand{\qdd}{\ddot{\q}}

\newcommand{\Psibar}{\greekvec{\Psi}}
\newcommand{\Psibardot}{\dot{\Psibar}}
\newcommand{\Psibarddot}{\ddot{\Psibar}}

\newcommand{\taubar}{\greekvec{\tau}}

\newcommand{\red}[1]{{\color{red}#1}}

\newcommand{\blue}[1]{{\color{blue}#1}}

\newcommand{\Phibar}{\boldsymbol{S}}

\usepackage{scalerel}
\newcommand{\timesM}{{\tilde{\smash[t]{\times}}}}
\newcommand{\timesfM}{\timesM {}^*}
\newcommand{\crffM}{\scaleobj{.87}{\,\tilde{\bar{\smash[t]{\!\times\!}}{}}}{}^{\,*}}

\usepackage{tikz}
\usetikzlibrary{matrix,calc}

\setlist[enumerate]{wide=0pt, widest=99,leftmargin=25pt, labelsep=*}

\newcommand{\J}{\boldsymbol{J}{}}
\newcommand{\K}{\boldsymbol{K}{}}
\renewcommand{\w}{\boldsymbol{w}}
\usepackage[makeroom]{cancel}
\newcommand{\Jc}{\J_{c}}
\newcommand{\Jdc}{\dot{\J_{c}}}

\renewcommand{\S}{\mathbf{S}\T}

\newcommand{\uM}{\mathbf{u}}
\newcommand{\XMD}{\mathbf{X}}
\newcommand{\lamM}{\boldsymbol{\lambda}}
\newcommand{\sM}{\mathbf{s}}
\newcommand{\SM}{\mathbf{S}}

\title{\Large \bf Multi-Shooting Differential Dynamic Programming for Hybrid Systems using Analytical Derivatives
}

\author{ Shubham Singh$^{1}$, Ryan P. Russell$^{1}$ and Patrick M.~Wensing$^{2}$
\thanks{$^{1}$ Aerospace Engineering, The University of Texas at Austin, TX-78751, USA. \href{mailto:singh281@utexas.edu}{singh281@utexas.edu}, \href{mailto:ryan.russell@utexas.edu}{ryan.russell@utexas.edu}}
       \thanks{$^{2}$ Aerospace \& Mechanical Engineering, University of Notre Dame, IN-46556, USA. \href{mailto:pwensing@nd.edu}{pwensing@nd.edu}
       } %
       \thanks{This work was supported in part by the National Science Foundation grants CMMI-1835013 and CMMI-1835186.}
       }

\begin{document}

\maketitle
\thispagestyle{empty}
\pagestyle{empty}


\begin{abstract}
Differential Dynamic Programming (DDP) is a popular technique used to generate motion for dynamic-legged robots in the recent past. However, in most cases, only the first-order partial derivatives of the underlying dynamics are used, resulting in the iLQR approach. Neglecting the second-order terms often slows down the convergence rate compared to full DDP. Multi-Shooting is another popular technique to improve robustness, especially if the dynamics are highly non-linear. In this work, we consider Multi-Shooting DDP for trajectory optimization of a bounding gait for a simplified quadruped model. As the main contribution, we develop Second-Order analytical partial derivatives of the rigid-body contact dynamics, extending our previous results for fixed/floating base models with multi-DoF joints. Finally, we show the benefits of a novel Quasi-Newton method for approximating second-order derivatives of the dynamics, leading to order-of-magnitude speedups in the
convergence compared to the full DDP method. 
\end{abstract}

\section{Introduction}
Trajectory optimization-based control has become a popular technique in the past few years for generating motion for legged robots like quadrupeds, humanoids, and manipulators~\cite{wensing2022optimization}. Many results in this area are based on using First-Order (FO) approximations of the rigid-body dynamics (RBD) in an approach called iLQR~\cite{tassa2012synthesis}. On the other hand, Differential Dynamic Programming (DDP)~\cite{mayne1966second} is a Second-Order (SO) method that has also gained varied interest for solving control problems in robotics~\cite{tassa2014control} \cite{li2020hybrid}. As compared to the super-linear convergence for iLQR, DDP offers quadratic convergence~\cite{murray1984differential} properties.

Multi-shooting versions of DDP and iLQR have also been proposed in the past for trajectory optimization in robotics. Giftthaler et al.~\cite{giftthaler2018family} proposed a family of multi-shooting iLQR algorithms and showed faster convergence over the single-shooting classical iLQR for problems involving a 6-DoF manipulator and the HyQ quadruped. Mastalli et al.~\cite{mastalli2020crocoddyl} 
 proposed an equivalent version of the Multiple-Shooting iLQR by a gap contraction strategy in the dynamics without adding extra decision variables to the problem. They also employ the FO partial derivatives of the contact dynamics (a.k.a.~KKT dynamics) from the popular RBD C\texttt{++} library Pinocchio~\cite{carpentier2019pinocchio}. MDDP~\cite{pellegrini2020multiple1}\cite{pellegrini2020multiple2} is a multi-shooting DDP optimizer for solving constrained optimal control problems (OCP) using the Augmented Lagrangian~\cite{bertsekas2014constrained} method to handle constraints. It was initially designed to solve OCPs relating to a variety of spacecraft trajectory applications. 

 For most of the works mentioned above, the partial derivatives of the cost and dynamics function, as needed for solving the OCP, are computed using either the finite numerical difference or the Automatic Differentiation (AD) approach. These methods are often approximate or slow to compute. However, analytical methods have also been used in the past to compute the FO partial derivatives of RBD~\cite{carpentier2018analytical,singh2022efficient,jain1993linearization}. In terms of computing the SO partial derivatives, there have been limited efforts. Lee et al.~\cite{lee2005newton} proposed a forward chain-rule approach to get the partial derivatives of the joint torque w.r.t.~joint configuration and joint rates for revolute and prismatic joint models. Nganga and Wensing~\cite{nganga2023accelerating} exploited reverse mode AD and a modified Recursive Newton Euler Algorithm (RNEA)~\cite{Featherstone08} to get the SO partial derivatives of the KKT dynamics without tensor computation. Although this strategy avoids the need for full SO partial derivatives, it lacks opportunities for parallel computation. As a previous contribution~\cite{singh2023second}, we presented the analytical recursive SO partial derivatives of Inverse and Forward unconstrained Dynamics using Featherstone's Spatial Vector Algebra (SVA)~\cite{Featherstone08} for models with multi-DoF joints. These SO partial derivatives are now extended in this work for systems with contacts, modeled using the KKT~\cite{udwadia1992new} dynamics.

\subsection{Contributions}
This paper presents two contributions to the use of multi-shooting DDP for rigid-body systems with contacts. The primary contribution is to provide analytical expressions for first-order and second-order partial derivatives of the KKT dynamics with respect to joint configuration, velocity, and joint torque (Sec~\ref{sec:ID_contacts}-\ref{sec:KKT_SO}). These derivatives extend the previously developed analytical derivatives by the authors. Using the newly developed analytical partial derivatives, an existing multi-shooting version of DDP (called MDDP) is employed to solve a trajectory optimization problem for a simplified 7-DoF quadruped (Sec.~\ref{sec:quad_problem}). Using this example, our secondary contribution is to explore the advantages of a Quasi-Newton approximation for second-order derivatives as compared to the full computation (Sec.~\ref{sec:QN_cases}).

\section{Rigid-Body Dynamics}
\label{sec:}
We consider a rigid-body system with state variables as the joint configuration $\q$ and joint velocity $\qd$, and the control variable as the joint torque vector $\taubar$. The unconstrained Inverse Dynamics (ID) for such a system is given as:
 \begin{equation}
    \S\taubar = \M(\q)\qdd+ \b(\q,\qd)
    = \textrm{ID}({\rm model},\q,  \qd, \qdd)
     \label{inv_dyn}
\end{equation}
The Coriolis and gravitational forces are combined in the term 
 $\b(\q,
 \qd)=\C(\q,\qd)\qd + \g(\q)$. The matrix $\mathbf{S}$ is the selector matrix to map the actuated joint torques to the entire tree. The matrix $\M \in \R^{n \times n}$ is the mass matrix, $\C \in \R^{n \times n}$ is the Coriolis matrix, $\g \in \R^{n}$ is the vector of generalized gravitational forces, and $n$ is the DoF of the system. The Recursive-Newton-Euler-Algorithm (RNEA)~\cite{Featherstone08} is a commonly used recursive algorithm used to compute $\taubar$ given $\q,\qd,\qdd$ in \eqref{inv_dyn}. The Forward Dynamics (FD) on the other hand computes $\qdd$ for a given $\taubar$:
\begin{equation}
    \qdd = \M^{-1}(\q)\left( \S\taubar -  \b(\q,\qd) \right)    = \textrm{FD}({\rm model},\q, \qd, \taubar)
     \label{fwd_dyn2}
\end{equation} 
For a system in contact, the dynamics are obtained using the KKT (Karush-Kuhn-Tucker) conditions and the Gauss principle of least constraint~\cite{udwadia1992new} as:
\begin{equation}
\underbrace{\left[\begin{array}{cc}
\M & \Jc^{\top} \\
\Jc & \mathbf{0}
\end{array}\right]}_{\K}\left[\begin{array}{c}
\qdd \\
-\lambda
\end{array}\right]=\left[\begin{array}{c}
\S \boldsymbol{\tau}-\b \\
-\Jdc\, \qd
\end{array}\right]
\label{KKT_eqn}
\end{equation}
which is simplified to get the so-called ``KKT-Dynamics'' as:
\begin{equation}
\left[\begin{array}{c}
\qdd \\
-\lambda
\end{array}\right]=\K^{-1} \left[\begin{array}{c}
\S \boldsymbol{\tau}-\b \\
-\Jdc\, \qd
\end{array}\right]
\label{KKT_dyn}
\end{equation}
Here, the matrix $\Jc \in \R^{n_{c} \times n}$ is the contact Jacobian for a contact mode with dimension $n_{c}$, and $\lambda \in \R^{n_{c}}$ is the vector of the contact forces. The augmented vector of joint accelerations and contact forces, $\a = [\qdd,\lambda]\T \in \R^{n+n_{c}}$ is computed by inverting the so-called KKT matrix~\cite{budhiraja2019multi}, denoted by $\K$ in \eqref{KKT_eqn}. This matrix inversion results in an $\mathcal{O}(N^3)$ operation, though lower-order algorithms are available~\cite{jain2012efficient}. Note that for un-constrained rigid-body systems, \eqref{KKT_dyn} reduces to \eqref{fwd_dyn2} and $\K=\M$. The first equation from \eqref{KKT_eqn} gives what we will call the ``constrained Inverse Dynamics'':
\begin{equation}
\begin{aligned}
     \S\taubar &= \M(\q)\qdd+ \b-\Jc\T \lambda \\
   &=  \textrm{ID}_c({\rm model},\q,  \qd, \qdd,\lambda)
     \end{aligned}
     \label{eqn_IDc}
\end{equation}
which is simply ID~\eqref{inv_dyn}, with the extra term from the contacts on the system. Impact events for the system follow impulsive Impact Dynamics, similar to the form in \eqref{KKT_eqn}, and given as:
\begin{equation}
\left[\begin{array}{c}
\qd_{+} \\
-\hat{\lambda}
\end{array}\right]=\K^{-1}\left[\begin{array}{c}
\M\qd_{-} \\
e \Jc \qd_{-}
\end{array}\right]
\label{im_eqn_1}
\end{equation}
Here, $e$ denotes the coefficient of restitution, set as 0 for perfectly inelastic collisions in our case. The impact event is assumed as impulsive, keeping the configuration $\q$ constant, while switching the joint velocities $\qd$. The quantities $\qd_{-}$ and $\qd_{+}$ denote the joint velocities pre and post-impact, respectively, while $\hat{\lambda} \in \R^{n_c}$ is the impulsive force vector acting on the body upon contact.

\section{Analytical Derivatives}
\label{sec:ana_derivs}
Previously, we presented the first/second-order analytical partial derivatives of the un-constrained Inverse/Forward Dynamics using Spatial Vector Algebra ~\cite{singh2022efficient,singh2022analytical,singh2023second}. Equation~\ref{jain_FO_eqn}\cite{jain1993linearization,carpentier2018analytical} give the relation between the FO un-constrained ID and FD partial derivatives w.r.t.~$\u =\{\q,\qd\}$, where $\u$ can be either $\q$ or $\qd$. An extension to compute the SO derivatives (Eq.~\eqref{ipr_so_1}) was presented in Ref.~\cite{singh2023second}. The SO FD derivatives in \eqref{ipr_so_1} also depend on the FO FD derivatives and SO ID derivatives. Here, the variable $\w=\{\q,\qd\}$, and the resulting SO derivative is a 3D tensor~\cite{singh2023second}. The operator $(\Rten)$~\cite{singh2023second} is a 3D tensor rotation of the elements of the tensor along the 2-3 dimension, which occurs following the product of $\frac{\partial \M}{\partial \u}$ and $\frac{\partial \textrm{FD}}{\partial \w}$. 
\begin{align}
      \scaleobj{1}{\frac{\partial\textrm{FD}}{\partial \boldsymbol u}} &\scaleobj{1}{= -\M^{-1}(\q) \frac{\partial\textrm{ID} }{\partial \boldsymbol u}}
      \label{jain_FO_eqn} \\
%
    \scaleobj{1
}{\frac{\partial^2 \textrm{FD} }{\partial \u \partial \w}} &\scaleobj{1}{= -\M^{-1}\Bigg( \frac{\partial^2 \textrm{ID}}{\partial \u \partial \w} + \frac{\partial \M}{\partial \w}\frac{\partial \textrm{FD}}{\partial \u} + \bigg(\frac{\partial \M}{\partial \u}\frac{\partial \textrm{FD}}{\partial \w}\bigg)\Rten \Bigg)}
    \label{ipr_so_1}
\end{align}

The following sections present an extension of the ID derivatives for un-constrained systems to the systems with external forces governed by \eqref{eqn_IDc}. Then the FO/SO derivatives of the KKT Dynamics~\eqref{KKT_eqn} w.r.t.~$\q,\qd,\taubar$ are presented. 
\subsection{Extension of ID Derivatives for Contacts:}
\label{sec:ID_contacts}
This section relies heavily on the use of Featherstone's Spatial Vector Algebra (SVA)~\cite[Ch.~2]{Featherstone08}. Section~II of our previous paper~\cite{singh2023second} provides a quick introduction to the topic. 
An open-chain kinematic tree is considered with $N$ links connected by $N$ joints, each up to 6 DoF. We define $i \preceq j$ if body $i$ is in the chain from body $j$ to the root. Joint $i$ is defined as connecting body $i$ and its predecessor~\cite{Featherstone08}. Spatial vectors are often represented either in a body-fixed frame or a ground-fixed frame. This section assumes that all the spatial vectors and derivatives are considered in the ground-fixed frame. In this paper, Euclidean vectors are denoted with lower-case letters with a bar ($\bar{v}$), spatial (6D) vectors \cite{Featherstone08} with lower-case bold letters (e.g., $\a$), and matrices with capitalized bold letters (e.g., $\boldsymbol{A}$). 
The $n$-vectors $\qd, \qdd, \taubar$ are also denoted by lower-case bold letters. \\

\subsubsection{FO Derivatives}
\label{sec:FO_contact}
We first extend the previously proposed result~\cite{singh2022efficient} on the FO partial derivatives of ID to handle systems with contacts. Using a similar approach presented in Ref.~\cite[Sec.~III]{singh2022efficient} to get the ID FO derivatives, we consider the joint torque for the joint $i$ for the constrained ID~\eqref{eqn_IDc} as:
\begin{equation}
    [\S\taubar]_{i} =   [ \M(\q) \qdd]_{i}+   [\C(\q,\qd) \qd]_{i}  + g_{i}(\q) - [\taubar_{ext}]_{i}
    \label{eqn_ID_con}
\end{equation}
 where $\taubar_{ext}=\Jc\T \lambda$ from \eqref{eqn_IDc}. From the RNEA~\cite{Featherstone08}, the  torque for joint $i$ is given via the spatial force across joint $i$:
\begin{equation}
    [\S\taubar]_{i} = \Phibar_{i}\T \sum_{l \succeq i}  ( \underbrace{\I_{l} \a_{l} + \v_{l} \times^{*} \I_{l} \v_{l}}_{\f_l} -\f_{l,ext}) 
        \label{eqn_RNEA_ext}
\end{equation}
where, $\I_{l}$, $\v_{l}$, $\a_{l}$, and $\f_{l,ext}$ are the spatial inertia, velocity, acceleration, and the external spatial force on the body $l$~\cite{Featherstone08}, all represented in a ground-fixed frame. The quantity $\Phibar_{i} \in \R^{6 \times n_{i}}$ is the Joint-Subspace-Motion matrix~\cite{Featherstone08} that maps the generalized joint velocity to the joint spatial velocity ($\v_{Ji} = \Phibar_{i}\qd_{i}$), while $n_{i}$ is the DoF of the joint.
Comparing~\eqref{eqn_ID_con} and \eqref{eqn_RNEA_ext}, $[\taubar_{ext}]_{i} =\Phibar_{i}\T \sum_{l \succeq i}\f_{l,ext}$.
The partial derivative of $[\taubar_{ext}]_{i}$ w.r.t.~$\q_{j}$ for $j \preceq i$ is now computed using some identities derived earlier in our previous work on the FO derivatives~\cite{singh2022efficient}. These identities give the partial derivative of basic kinematic and dynamic quantities w.r.t.~$\q$ and $\qd$. Using the identity J3~\cite[Appendix]{singh2022efficient}:
\begin{equation}
    \frac{\partial [\taubar_{ext}]_{i}}{\partial \q_{j}} =-\Phibar_{i}\T (\f_{i,ext}^{C} \crff  \Phibar_{j})+ \Phibar_{i}\T\frac{\partial \f_{i,ext}^{C}}{\partial \q_{j}}, (j \preceq i)
   \label{JTlam_i_term_1}
\end{equation}
where, $\f_{i,ext}^{C}=\sum_{l \succeq i}\f_{l,ext}$ is the cumulative external spatial force on body $i$. The derivative of $[\taubar_{ext}]_j$ w.r.t.~$\q_{i}$ is:
\begin{equation}
    \frac{\partial [\taubar_{ext}]_{j}}{\partial \q_{i}} =\Phibar_{j}\T\frac{\partial \f_{j,ext}^{C}}{\partial \q_{i}}, (j \prec i)
   \label{JTlam_i_term_2}
\end{equation}
Hence, the updated ID FO derivatives~\cite[Eq.~27]{singh2022efficient}  now are:
\begin{equation}
\begin{aligned}
    & \frac{\partial \taubar_{i}}{\partial \q_{j}} =  \Phibar_{i} \T \big[ 2  \B_{i}^{C} \big]\Psibardot{}_{j} + \Phibar_{i} \T \I_{i}^{C} \Psibarddot{}_{j}  +\\
&~~~~~~~~~~~~~~~~~~~~\Phibar_{i}\T (\f_{i,ext}^{C} \crff  \Phibar_{j}) -\Phibar_{i}\T\blue{\frac{\partial \f_{i,ext}^{C}}{\partial \q_j}}, (j \preceq i) \\
    & \frac{\partial \taubar_{j}}{\partial \q_{i}} = \Phibar_{j}\T [ 2 \B_{i}^{C} \Psibardot_{i} +  \I_{i}^{C}  \Psibarddot_{i}+(\f_{i}^{C} )\crff \Phibar_{i}  ]-\\
    &~~~~~~~~~~~~~~~~~~~~~~~~~~~~\Phibar_{j}\T \red{\frac{\partial \f_{j,ext}^{C}}{\partial \q_{i}}} , (j \prec i)  
\end{aligned}
\label{eqn_ID_FO_contact}
\end{equation}
Equations~\eqref{eqn_ID_FO_contact} denote the FO Lie-derivatives of $\taubar$, as were defined in Ref.~\cite{singh2022efficient}. Here, $\Psibardot$ and $\Psibarddot$ were defined in Ref.~\cite[Eq.~28]{singh2022efficient}, the quantity $\B_{i}^{C}$ is a body-level Coriolis matrix~\cite{echeandia2021numerical}, and $\f_{i}^{C}=\sum_{l \succeq i}\f_{l}$ is the cumulative spatial force. The partial derivatives of constrained ID~\eqref{eqn_IDc} w.r.t.~$\qd$ remain unchanged since the quantity $\Jc\T \lambda$ is only a function of configuration. 

 For getting the generalized joint torque from \eqref{eqn_IDc}, and its derivative, the contact force $\lambda$ is assumed as a constant input.
 The quantity $\f_{i,ext}^{C}$ and its partial derivative w.r.t.~$\q$ depends on $\lambda$ and the geometry of the contact point on the connectivity tree.
To explain, consider a single point contact on link $m$. (Other contact configurations can always be represented via multiple points).  Then, the external spatial force on link $m$ is:
\begin{equation}
    \f_{m,ext} = \left[\begin{array}{c} {}^{0}{\bar{p}_{c}} \times {}^{0}\bar{\lambda}_{c} \\ {}^{0}\bar{\lambda}_{c} \end{array}\right]
    \label{fcext_eqn}
\end{equation}
where, ${}^{0}\bar{\lambda}_{c} \in \R^3$ is the 3D Cartesian force on link $m$,  while ${}^{0}{\bar{p}_{c}}$ is the position vector of the point contact, both represented in the ground-fixed frame. The quantity $ {}^{0}{\bar{p}_{c}} \times {}^{0}\bar{\lambda}_{c}$ in \eqref{fcext_eqn} is the moment due to $\lambda$ about the origin of the ground frame. The partial derivative of $\f_{m,ext}$ w.r.t.~$\q_{j}$, where $j\preceq m$ is:
\begin{equation}
   \frac{\partial \f_{m,ext}}{\partial \q_{j}} =  \left[\begin{array}{c} -{}^{0}\bar{\lambda}_{c}\times \Jc(:,j) \\ \mathbf{0}_{3 \times n_j} \end{array}\right]
\end{equation}
where, $\Jc(:,j)$ represents the columns of the contact Jacobian $\Jc$ corresponding to joint $j$. For a single contact at the body $m$, $\f_{i,ext}^{C}=\f_{m,ext}$, if $i \preceq m$, and zero, otherwise. Hence, the partial derivative of $\f_{i,ext}^{C}$ w.r.t.~$\q_{j}$ is:
\begin{align}
  \blue{\frac{\partial \f_{i,ext}^{C}}{\partial \q_{j}}}=
    \begin{cases}
       \left[\begin{array}{c} -{}^{0}\bar{\lambda}_{c}\times \Jc(:,j) \\ \mathbf{0}_{3 \times n_j} \end{array}\right] ,  &~~~~ \text{if}\  j \preceq i \preceq m \\
      \mathbf{0}_{6 \times n_j},  &~~~~ \text{otherwise}
    \end{cases}
    \label{eqn_fic_ext_FO_1}
\end{align}
Similar algebra can be done for the case $j\succ i$, to get the term $\frac{\partial \f_{j,ext}^{C}}{\partial \q_{i}}$.

The IDSVA algorithm~\cite{singh2022efficient} is now updated with the new terms to return the FO partial derivatives of ID~\eqref{eqn_ID_FO_contact}. An implementation of related contact derivatives can also be found in the C\texttt{++} Pinocchio~\cite{carpentier2019pinocchio} open-source library.
\subsubsection{SO Derivatives}
\label{sec:SO_contact}
The ID-SO partial derivatives~\cite{singh2023second} of \eqref{inv_dyn} are also extended for systems with external forces. The only extra term needed is the SO partial derivative of the term $[\taubar_{ext}]$ in \eqref{eqn_ID_con}. The details of the derivation are skipped here and can be found in Ref.~\cite[Sec.~X]{arxivss_SO}. However, the final expressions are provided here. Following the approach in Ref.~\cite{singh2023second}, another FO partial derivative of \eqref{JTlam_i_term_1} is taken w.r.t.~$\q_{k}$ to obtain:
\begin{equation}
\begin{aligned}
    &\frac{\partial^2 (\taubar_{ext})_{i}}{\partial \q_{j} \partial \q_{k}} = \Phibar_{i}\T (\f_{i,ext}^{C} \crff \Phibar_{k} )\crffM \Phibar_{j} - \\
    &\Phibar_{i}\T\Bigg[\bigg(\frac{\partial \f_{i,ext}^{C}}{\partial \q_{k}}\crffM\bigg) \Phibar_{j}+ \Phibar_{k} \timesfM \frac{\partial \f_{i,ext}^{C}}{\partial \q_{j}}  \Bigg] + \Phibar_{i}\T \frac{\partial^2 \f_{i,ext}^{C}}{\partial \q_{j} \partial \q_{k}}
\end{aligned}
    \label{ID_SO_contact_1}
\end{equation}
which hold when $k \preceq j \preceq i$. 
Similarly, the other derivatives under conditions $(k \preceq j \prec i)$ and $(k \prec j \preceq i)$ resp.~are:
\begin{align}
    \scaleobj{1.05}{\frac{\partial^2 (\taubar_{ext})_{j}}{\partial \q_{k} \partial \q_{i}}} &\scaleobj{1.05}{= -\Phibar_{j}\T\Bigg[\bigg(\frac{\partial  \f_{j,ext}^{C}}{\partial \q_{i}}\crffM\bigg) \Phibar_{k}  \Bigg] + \Phibar_{j}\T \frac{\partial^2 \f_{j,ext}^{C}}{\partial \q_{k} \partial \q_{i}}}
        \label{ID_SO_contact_2} \\
    \scaleobj{1.05}{\frac{\partial^2 (\taubar_{ext})_{k}}{\partial \q_{i} \partial \q_{j}}} &\scaleobj{1.05}{= \Phibar_{k}\T \frac{\partial^2 \f_{k,ext}^{C}}{\partial \q_{i} \partial \q_{j}}}
        \label{ID_SO_contact_3}
\end{align}

Many tensorial SVA identities and properties from Ref.~\cite[App-B,Table-II]{singh2023second} are used to get these expressions.
 The SO derivative $\frac{\partial^2 \f_{ext}^{C}}{\partial \q^2}$ is required in \eqref{ID_SO_contact_1}-\eqref{ID_SO_contact_3} and can be computed similar to the FO derivatives, building from \eqref{eqn_fic_ext_FO_1}. The IDSVA-SO algorithm~\cite{singh2023second} is modified to return the SO partial derivatives of ID with contacts. The details of those are skipped here for brevity but can be found in Ref.~\cite[Sec.~X]{arxivss_SO}.












\subsection{KKT Dynamics FO Derivatives}
\label{sec:KKT_FO}
For the optimization problem formulated in this paper, we need the partial derivatives of the KKT Forward Dynamics function ($\a=[\qdd,\lambda]\T$) w.r.t.~the joint states $\q,\qd$ and the control variable $\taubar$. We assume $\u=\{\q,\qd,\taubar \}$ and take the FO partial derivative of \eqref{KKT_eqn} w.r.t.~$\u$ to get the set of equations:
\begin{equation}
\begin{aligned}
       \frac{\partial \M}{\partial \u}\qdd+\M\frac{\partial \qdd}{\partial \u} - \frac{\partial \Jc\T}{\partial \u}\lambda - \Jc\T \frac{\partial \lambda}{\partial \u} &=  \S\frac{\partial \taubar}{\partial \u}-\frac{\partial \b}{\partial \u}\\
    \frac{\partial \Jc}{\partial \u}\qdd + \Jc\frac{\partial \qdd}{\partial \u} + \frac{\partial (\Jdc\qd)}{\partial \u} &=0
\end{aligned}
    \label{FO_crude_eqn}
\end{equation}
Upon simplification, we get the following set of equations for the FO partial derivative of KKT FD~\cite[Eq.~3.16]{budhiraja2019multi}:
\begin{equation}
\resizebox{1.0\hsize}{!}{$
\left[\begin{array}{c}
\frac{\partial \qdd}{\partial \u} \\[1ex]
-\frac{\partial \lambda}{\partial \u}
\end{array}\right] = -\left[\begin{array}{cc}
\M & \Jc^{\top} \\
\Jc & \mathbf{0}
\end{array}\right]^{-1}\left[\begin{array}{c}
\frac{\partial \M}{\partial \u}\qdd+\frac{\partial \b}{\partial \u}-\frac{\partial \Jc\T}{\partial \u}\lambda-\S\frac{\partial \taubar}{\partial \u}\\[1ex]
\frac{\partial \Jc}{\partial \u}\qdd+\frac{\partial (\Jdc\qd)}{\partial \u}
\end{array}\right]$}
\label{FO_general_eqn}
\end{equation}
The dimension of the derivatives from \eqref{FO_general_eqn} is $(n+n_{c})\times n$. 
For $\u=\q$:
\begin{equation}
\resizebox{1.0\hsize}{!}{$
\left[\begin{array}{c}
\frac{\partial \qdd}{\partial \q} \\[1ex]
-\frac{\partial \lambda}{\partial \q}
\end{array}\right] = -\left[\begin{array}{cc}
\M & \Jc^{\top} \\
\Jc & \mathbf{0}
\end{array}\right]^{-1}\left[\begin{array}{c}
\frac{\partial (\M\qdd+\b-\Jc\T\lambda)}{\partial \q}\\
\frac{\partial \Jc}{\partial \q}\qdd+\frac{\partial (\Jdc\qd)}{\partial \q}
\end{array}\right]$}
\label{ID_FO_q_eqn1}
\end{equation}
where the last term $\S\frac{\partial \taubar}{\partial \q}$ is zero, since $\taubar$ is an input, and not dependent on $\q$. Equation~\eqref{eqn_ID_FO_contact} is used to compute the $\textrm{ID}_{c}$ FO derivative embedded in \eqref{ID_FO_q_eqn1}. Using the notation ($\K$) for the KKT matrix:
\begin{equation}
\left[\begin{array}{c}
\frac{\partial \qdd}{\partial \q} \\[1ex]
-\frac{\partial \lambda}{\partial \q}
\end{array}\right] = -\K^{-1}\left[\begin{array}{c}
\frac{\partial \textrm{ID}_{c}}{\partial \q}\\
\frac{\partial (\Jc \qdd+\Jdc\qd)}{\partial \q}
\end{array}\right]\Bigg|_{\q_0, \qd_0,\qdd_{0},\lambda_{0}}
\label{ID_FO_q_eqn2}
\end{equation}
%
The KKT FD from \eqref{KKT_dyn} is first evaluated for any state $\q_{0},\qd_{0}$, and control $\taubar_{0}$ to get $\qdd_{0}$ and $\lambda_{0}$, needed for \eqref{ID_FO_q_eqn2}. The partial derivatives of the contact acceleration $\Jc\qdd+\Jdc\qd$ required in \eqref{ID_FO_q_eqn2} are computed separately by taking the chain rule derivative of $\Jc\qdd+\Jdc\qd$ w.r.t.~$\q$. This calculation is straightforward since a rigid-body system can be represented using a connectivity tree. The position vector of the contact point is purely a function of $\q$. Taking the partial derivative of this vector in the ground-fixed frame will give us $\Jc$, represented in the ground frame. 

 Using $\u=\qd$ in \eqref{FO_general_eqn} results to:
\begin{equation}
\left[\begin{array}{c}
\frac{\partial \qdd}{\partial \qd} \\
-\frac{\partial \lambda}{\partial \qd}
\end{array}\right] = -\K^{-1}\left[\begin{array}{c}
\frac{\partial \textrm{ID}_c}{\partial \qd}\\
\frac{\partial (\dot{\J_{c}}\qd)}{\partial \qd}
\end{array}\right]\Bigg|_{\q_0, \qd_0,\qdd_{0},\lambda_{0}}
\label{ID_FO_v_eqn1}
\end{equation}
where partial derivative of $\Jdc\qd$ w.r.t.~$\qd$ is computed separately using a chain-rule derivative approach. Similar analysis can also be done for $\u=\taubar$ in \eqref{FO_general_eqn}.

\subsection{KKT Dynamics SO Derivatives}
\label{sec:KKT_SO}
To get the SO derivatives of the KKT FD \eqref{KKT_eqn}, we take another partial derivative of FO partial derivatives derived in the previous section. Taking the FO partial derivative of \eqref{FO_crude_eqn} w.r.t.~$\w$ as:

\begin{equation}
\resizebox{1.0\hsize}{!}{$
    \begin{aligned}
      \frac{\partial^2 \M}{\partial \u \partial \w}\qdd+\frac{\partial \M}{\partial \w}\frac{\partial \qdd}{\partial \u} +\Bigg( \frac{\partial \M}{\partial \u}\frac{\partial \qdd}{\partial \w}\Bigg) \Rten +\M\frac{\partial^2 \qdd}{\partial \u\partial \w}+\frac{\partial^2 \b}{\partial \u \partial \w} =  \\ \frac{\partial \Jc\T}{\partial \w} \frac{\partial \lambda}{\partial \u}+\Jc\T\frac{\partial ^2 \lambda}{\partial \u \partial \w}+\frac{\partial^2 \Jc\T}{\partial \u \partial \w}\lambda +\Bigg(\frac{\partial \Jc\T}{\partial \u} \frac{\partial \lambda}{\partial \w}\Bigg)\Rten +\S\frac{\partial^2 \taubar}{\partial \u \partial \w}  
\end{aligned}$}
\label{eqn_KKT_SO_1}
\end{equation}
\begin{equation}
       \Bigg(\frac{\partial \Jc}{\partial \u}\frac{\partial \qdd}{\partial \w}\Bigg)\Rten+\frac{\partial \Jc}{\partial \w}\frac{\partial \qdd}{\partial \u}+\Jc\frac{\partial^2 \qdd}{\partial \u \partial \w} =-\frac{\partial^2 \Jc}{\partial \u \partial \w}\qdd -\frac{\partial^2 (\Jdc\qd)}{\partial \u \partial \w}
        \label{eqn_KKT_SO_2}
\end{equation}
The product of tensor and matrix follow the rules defined for the tensorial SVA in Ref.~\cite{singh2023second}. The 2-3 rotations $(\Rten)$ on the tensor terms in \eqref{eqn_KKT_SO_1},\eqref{eqn_KKT_SO_2} are due to the underlying tensor storage order assumptions. To explain this order, we look at the term $\frac{\partial^2 \b}{\partial \u \partial \w}$ in \eqref{eqn_KKT_SO_1}. Here, the elements of $\b$ are along the rows of the 3D tensor, $\u$ along the columns, and $\w$ are along the pages of the tensor. However, the tensor-matrix product term $\frac{\partial \M}{\partial \u}\frac{\partial \qdd}{\partial \w}$ would result in a term  with elements of $\u$ along the pages, and $\w$ along the columns. Hence, based on the assumed tensor storage order, this term is rotated in the 2-3 dimension for consistency. This manipulation of terms can also be understood by assuming $\u$ as a vector and $\w$ as a scalar. In that case, the 2-3 rotation would ``flatten'' the matrix from the 2-3 dimension to the 1-2 dimension. Simplifying the equations above:
\begin{equation}
\resizebox{1.0\hsize}{!}{$
\begin{aligned}
    \left[\begin{array}{cc}
    \M & \Jc^{\top} \\[2ex]
    \Jc & \mathbf{0}
    \end{array}\right]\left[\begin{array}{c}
    \frac{\partial^2 \qdd}{\partial \u \partial \w} \\[2ex]
    \frac{\partial^2 (-\lambda)}{\partial \u \partial \w}
    \end{array}\right]+   \left[\begin{array}{cccc}
    \frac{\partial \M}{\partial \u} & \frac{\partial \M}{\partial \w} & \frac{\partial \Jc\T}{\partial \u} & \frac{\partial \Jc\T}{\partial \w}\\[2ex]
    \frac{\partial \Jc}{\partial \u} & \frac{\partial \Jc}{\partial \w}& \mathbf{0}& \mathbf{0}
    \end{array}\right] \left[\begin{array}{c}
    \frac{\partial \qdd}{\partial \w} \\ [.5ex]
    \frac{\partial \qdd}{\partial \u} \\[.5ex]
    -\frac{\partial \lambda}{\partial \w} \\[.5ex]
    -\frac{\partial \lambda}{\partial \u}
    \end{array}\right] = \\
    -\left[\begin{array}{c}
    \frac{\partial^2 \M}{\partial \u \partial \w}\qdd+\frac{\partial^2 \b}{\partial \u \partial \w}-\frac{\partial^2 \Jc\T}{\partial \u \partial \w}\lambda  -\S\frac{\partial^2 \taubar}{\partial \u \partial \w}\\[2ex]
    \frac{\partial^2 \Jc}{\partial \u \partial \w}\qdd +\frac{\partial^2 (\Jdc\qd)}{\partial \u \partial \w}
    \end{array}\right]
    \end{aligned}
$}
\label{SO_general_eqn}
\end{equation}

Equation \eqref{SO_general_eqn} is an extension of \eqref{FO_general_eqn} for the SO case. The following sections give the expressions of SO derivatives for different cases by selecting $\u,\w$ as $\q$, $\qd$, and $\taubar$. 
\subsubsection{Derivatives w.r.t.~$\q$:}
For $\u,\w=\q$:
\begin{equation}
\resizebox{1.0\hsize}{!}{$
\begin{aligned}
   &  \left[\begin{array}{c}
    \frac{\partial^2 \qdd}{\partial \q^2 } \\
    \frac{\partial^2 (-\lambda)}{\partial \q^2 }
    \end{array}\right]   = -\K^{-1}
    \left[\begin{array}{c}
    \frac{\partial^2 \textrm{ID}_c}{\partial \q^2}+ \\
     \frac{\partial^2 (\Jc \qdd+\Jdc\qd)}{\partial \q^2 } +\end{array} \right. \\
  & \left.\begin{array}{c} \frac{\partial \M}{\partial \q}\frac{\partial \qdd}{\partial \q}+\Big(\frac{\partial \M}{\partial \q}\frac{\partial \qdd}{\partial \q} \Big)\Rten -  \frac{\partial \Jc\T}{\partial \q}\frac{\partial \lambda}{\partial \q} -\Big( \frac{\partial \Jc\T}{\partial \q}\frac{\partial \lambda}{\partial \q} \Big)\Rten \\
   \frac{\partial \Jc}{\partial \q}\frac{\partial \qdd}{\partial \q}+  \Big(\frac{\partial \Jc}{\partial \q}\frac{\partial \qdd}{\partial \q}\Big)\Rten
    \end{array}    
     \right]\Bigg|_{\q_0, \qd_0,\qdd_{0},\lambda_{0}}
 \end{aligned}$}
 \label{eqn_KKT_SO_q}
 \end{equation}
For a given $\q_{0}$, $\qd_{0}$, and $\taubar_{0}$, $\qdd_{0}$, and $\lambda_{0}$ are computed using \eqref{KKT_dyn}. The FD FO partial derivatives required in~\eqref{eqn_KKT_SO_q} are computed using \eqref{ID_FO_q_eqn2}. The SO derivative of constrained ID, $\frac{\partial^2 \textrm{ID}_{c}}{\partial \q^2}$, and $\frac{\partial \M}{\partial \q}$ in \eqref{eqn_KKT_SO_q} are computed using a modified version of IDSVA-SO algorithm from Ref.~\cite{singh2023second}, that employs \eqref{ID_SO_contact_1}-\eqref{ID_SO_contact_3}. Similar to the FO case, the SO partial derivative of the constraint $\Jc\qdd+\Jdc\qd$ w.r.t. $\q$ is computed using a chain-rule approach done manually. The tensor terms $\frac{\partial \Jc}{\partial \q} \in \R^{n_{c}\times n \times n}$ and $\frac{\partial \Jc\T}{\partial \q} \in \R^{n \times n_{c} \times n}$ are also computed manually in the ground frame.
 \subsubsection{Derivatives w.r.t.~$\qd$:}
 For $\u,\w = \qd$, \eqref{SO_general_eqn} reduces to:
 \begin{equation}
     \begin{aligned}
    \left[\begin{array}{c}
    \frac{\partial^2 \qdd}{\partial \qd^2 } \\[.75ex]
    \frac{\partial^2 (-\lambda)}{\partial \qd^2 }
    \end{array}\right]   = -\K^{-1}\left[\begin{array}{c}
    \frac{\partial^2 \textrm{ID}_{c}}{\partial \qd^2} \\[.75ex]
    \frac{\partial^2 (\Jdc\qd)}{\partial \qd^2}\end{array}\right]\Bigg|_{\q_0, \qd_0,\qdd_{0},\lambda_{0}}
    \end{aligned}
     \label{eqn_KKT_SO_v}
\end{equation}
 \subsubsection{Derivatives w.r.t.~$\qd$,$\q$:}
 For $\u=\qd$, and $\w = \q$, \eqref{SO_general_eqn} reduces to
\begin{equation}
\resizebox{1.0\hsize}{!}{$
     \begin{aligned}
    \left[\begin{array}{c}
    \frac{\partial^2 \qdd}{\partial \qd \partial \q} \\[.75ex]
    \frac{\partial^2 (-\lambda)}{\partial \qd \partial \q}
    \end{array}\right]    = -\K^{-1}\left[\begin{array}{c}
    \frac{\partial^2 \textrm{ID}_{c}}{\partial \qd \partial \q}+ \frac{\partial \M}{\partial \q}\frac{\partial \qdd}{\partial \qd} - \frac{\partial \Jc\T}{\partial \q}\frac{\partial \lambda}{\partial \qd}\\[.75ex]
    \frac{\partial^2 (\Jdc\qd)}{\partial \qd \partial \q}+\frac{\partial \Jc}{\partial \q}\frac{\partial \qdd}{\partial \qd}
    \end{array}\right]\Bigg|_{\q_0, \qd_0,\qdd_{0},\lambda_{0}}
    \end{aligned}$}
         \label{eqn_KKT_SO_vq}
\end{equation}
 The SO partial derivative of $(\Jdc\qd)$ w.r.t.~$\qd$, and the cross-SO derivative w.r.t. $\qd,\q$ are computed manually, and used in \eqref{eqn_KKT_SO_v}, and \eqref{eqn_KKT_SO_vq} respectively.

To get the cross-derivative where $\u=\q$ and $\w=\qd$, we 
 use the symmetry property of the Hessian as:
\begin{equation}
      \left[\begin{array}{c}
    \frac{\partial^2 \qdd}{\partial \q \partial \qd} \\[.75ex]
    \frac{\partial^2 (-\lambda)}{\partial \q \partial \qd}
    \end{array}\right]  =  \Bigg( \left[\begin{array}{c}
    \frac{\partial^2 \qdd}{\partial \qd \partial \q} \\[.75ex]
    \frac{\partial^2 (-\lambda)}{\partial \qd \partial \q}
    \end{array}\right]  \Bigg)\Rten
\end{equation}
This property was also exploited in Ref.~\cite[Sec.~IV E]{singh2023second} to get the cross-derivatives of ID w.r.t.~$\q$,$\qd$.
\subsubsection{Derivatives w.r.t.~$\taubar$, $\q$:}
For $\u=\taubar$, $\w = \q$, \eqref{SO_general_eqn} reduces to:
\begin{align}
\left[\begin{array}{c}
\frac{\partial^2 \qdd}{\partial \taubar \partial \q} \\
\frac{\partial^2 (-\lambda)}{\partial \taubar \partial \q}
\end{array}\right]     = -\K^{-1}\left[\begin{array}{c}
 \frac{\partial \M}{\partial \q} \frac{\partial \qdd}{\partial \taubar}-\frac{\partial \Jc\T}{\partial \q}\frac{\partial \lambda}{\partial \taubar}\\
 \frac{\partial \Jc}{\partial \q}\frac{\partial \qdd}{\partial \taubar}\end{array}\right]\Bigg|_{\q_0, \qd_0,\qdd_{0},\lambda_{0}}
\end{align}

Using \eqref{KKT_eqn}, the cases, $\frac{\partial^2 \a}{\partial \qd \partial \taubar}$, and $\frac{\partial^2 \a}{\partial^2 \taubar}$ results in zero partial derivatives. This was also the case for SO partial derivatives of FD for un-constrained systems~\eqref{fwd_dyn2}, as shown in Ref.~\cite{singh2022efficient}. 

To get the partial derivatives of the Impact dynamics~\eqref{im_eqn_1}, a similar process from Sec.~\ref{sec:KKT_FO},\ref{sec:KKT_SO} is followed. Here, the derivatives are computed w.r.t.~$\u,\w=\{\q,\qd_{-}\}$. The details of the derivation are skipped here and can be found in Ref.~\cite[Sec.~XII]{arxivss_SO}, but the final form of the equations is similar to Eq.~\eqref{FO_general_eqn} and \eqref{SO_general_eqn}.

 The FO/SO analytical partial derivatives of the KKT and the impact dynamics were compared against the complex-step method~\cite{squire1998using,lantoine2012using} for accuracy verifications. A 7-DoF planar 2D quadruped model of the MIT Mini-Cheetah~\cite{katz2019mini} is used for benchmarking the presented derivatives. As an extra accuracy check, the CasADi toolbox~\cite{andersson2019casadi} implemented in MATLAB was also used to verify the analytical partial derivatives. An open-source version of these derivatives integrated with Featherstone's~\cite{Featherstone08} \texttt{spatial\_v2} library is provided at Ref.~\cite{matlabsource}.

\section{Multi-Shooting DDP}
\label{sec:MDDP}
\begin{figure}[t]
 \centering
 \includegraphics[width=0.85 \columnwidth]{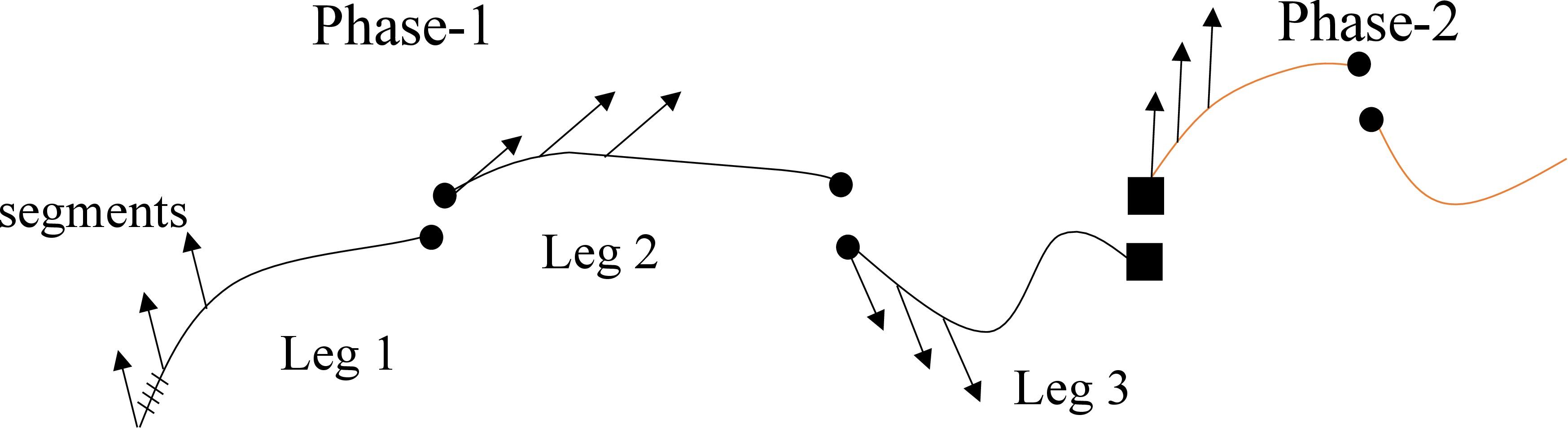}
 \caption{Multi-shooting leg/phase structure in MDDP~\cite[Fig.~2]{pellegrini2020multiple1}. Each leg consists of segments over which the controls are constant. Each segment is divided into steps (shown for the first segment of first leg) for numerical integration. Inter-leg and inter-phase constraints are denoted by filled circles and squares respectively.}
 \label{MDDP_legs}
 \end{figure}

The optimization problem formulated in this paper is solved using the Multiple Shooting Differential Dynamic Programming (MDDP) algorithm proposed by Pelligrini and Russell~\cite{pellegrini2020multiple1,pellegrini2020multiple2,pellegrini2012quasi}. MDDP is an extension of the fundamental inner loop from Ref.~\cite{lantoine2012hybrid1} where the computation of the dynamics and optimization of the
classic DDP algorithm was decoupled to allow for parallel computation and quasi-newton approximations of the dynamics only, among other benefits. The software implementing the MDDP algorithm allows the user to set up a multi-phase problem, where each ``phase'' has a different dynamics, cost, terminal, and path constraints function associated with it. The different phases are connected to each other using ``linkage'' constraints, which ensures continuity across the phase boundary. The user can also select which states have to be linked, or set free for a phase. Each phase can also be split into multiple sub-intervals called ``legs'', which are also connected using inter-leg linkage constraints (Fig.~\ref{MDDP_legs}). The MDDP algorithm exploits this multi-leg and multi-phase feature to optimize for the initial conditions of each leg, thus giving rise to its multiple-shooting feature. The MDDP formulation includes cross-leg feedback terms during the forward sweep that are not present in single shooting DDP~\cite{pellegrini2020multiple1}. MDDP allows the user to select one or more legs for a particular phase. For the current study, each phase has been assumed to have a single ``leg'' for convenience but can be changed as a tuning parameter to reduce the sensitivity of the problem.

The MDDP algorithm employs multiple features including the Augmented Lagrangian~\cite{bertsekas2014constrained} method to handle the terminal and path constraints. The inner-loop~\cite[Sec.~5]{pellegrini2020multiple1} in MDDP solves for the un-constrained problem, followed by the outer-loop where the Lagrange multipliers are updated for the path and terminal constraints. Another important feature of MDDP is the Trust-Region (TR) method~\cite{levenberg1944method} to solve the quadratic sub-problems formulated in the backward pass of DDP. The control laws~\cite[Eq.~61,88,97]{pellegrini2020multiple1} developed using the TR method are:
\begin{equation}
\begin{aligned}
\delta \mathbf{\uM}_k&=-\tilde{J}_{\uM \uM, k}^{-1}\left(J_{\uM, k}+J_{\uM \XMD, k} \delta \mathbf{\XMD}_k+J_{\uM \lamM, k} \delta \lamM+J_{\uM \sM} \delta \sM\right) \\
\delta \sM&=-\tilde{J}_{\SM \SM}^{-1}\left(J_\SM+J_{\SM \Lambda} \delta \mathbf{\Lambda}\right) \\ 
\delta \boldsymbol{\Lambda}&=\tilde{\hat{J}}_{\Lambda \Lambda}^{-1} \hat{J}_{\Lambda}
\end{aligned}
\label{eqn_control}
\end{equation}
Here, $\delta \uM_k$ is the control law for the $k^{th}$ time-step in the backward pass, $J$ is the cost-to-go for the problem, $\XMD$ is the state vector, $\lamM$ is the vector of Lagrange multipliers for a particular leg, $\sM$ forms the vector of the initial states of a leg, $\SM$ is the augmented vector of $\sM$ for all the legs, and $\mathbf{\Lambda}$ is the augmented vector of $\lamM$ for all the legs. Note the re-use of the symbol $\lambda$, which was earlier denoted as the contact force on a rigid-body system. The terms on the right-hand side of \eqref{eqn_control} comprise of the partial derivatives. Unlike the classical DDP approach, these partial derivatives of $J$ are propagated in the backward pass using the State-Transition-Matrix (STM)~\cite{junkins2008high}, which is computed initially in the forward pass. As an example, the partial derivatives of $J$ w.r.t.~the state ($\XMD$) and the control ($\uM$) are mapped in the backward pass as~\cite[Eq.~50]{pellegrini2020multiple1}:
\begin{equation}
\left[\begin{array}{c}
J_{\XMD, k} \\
J_{\uM, k}
\end{array}\right]\T=\left[\begin{array}{c}
L_{\XMD, k} \\
L_{\uM, k}
\end{array}\right]\T+\left[\begin{array}{c}
J_{\XMD, k+1}^* \\
\mathbf{0}_{n_\uM}
\end{array}\right]\T \Phi_k^1
\label{eqn_STM}
\end{equation}
where $J_{\XMD, k+1}^*$ denotes the partial derivative of $J$ w.r.t.~$\XMD$ evaluated at the future optimized step $k+1$,  $L$ is the running cost, and $\Phi_k^1$ is the FO STM for the $k^{th}$ step, computed in the forward pass using propagation as $\dot{\Phi}^1=f_\XMD \Phi^1$. Here $f_\XMD$ is the FO partial of the dynamics function, which resolves to the FD~\eqref{fwd_dyn2} and KKT FD \eqref{KKT_dyn} for our case. A similar SO STM $\Phi^{2}$ is also used to map the SO partial derivatives of the cost-to-go in the backward pass, and the details can be found in Ref.~\cite[Eq.~51]{pellegrini2020multiple1}.

Another useful feature implemented in MDDP is the Quasi-Newton (QN) 
approximation~\cite{fletcher1963rapidly} method to compute the SO STM. Historically, applied to DDP or other optimization solvers, QN methods are used to approximate the Hessian of the cost function directly with
respect to the controls~\cite{sen1987quasi}. Instead, the QN method here approximates the SO STM only, while the SO cost to go partial computations are unaffected. The FO STM is used along with the states at one iteration to estimate the SO STM for the next iteration during the forward pass of the DDP.  Every p$^{th}$ iteration, this QN approximation (referred to as QNp) is replaced with the full SO dynamics, where the frequency p of the full update is a
tuning parameter. When p is large, the bulk of the iterations are evaluated with FO derivatives only.

\section{Optimization Problem for Simplified Quadruped}
\label{sec:quad_problem}

\begin{figure}[t]
 \centering
 \includegraphics[width=0.75 \columnwidth]{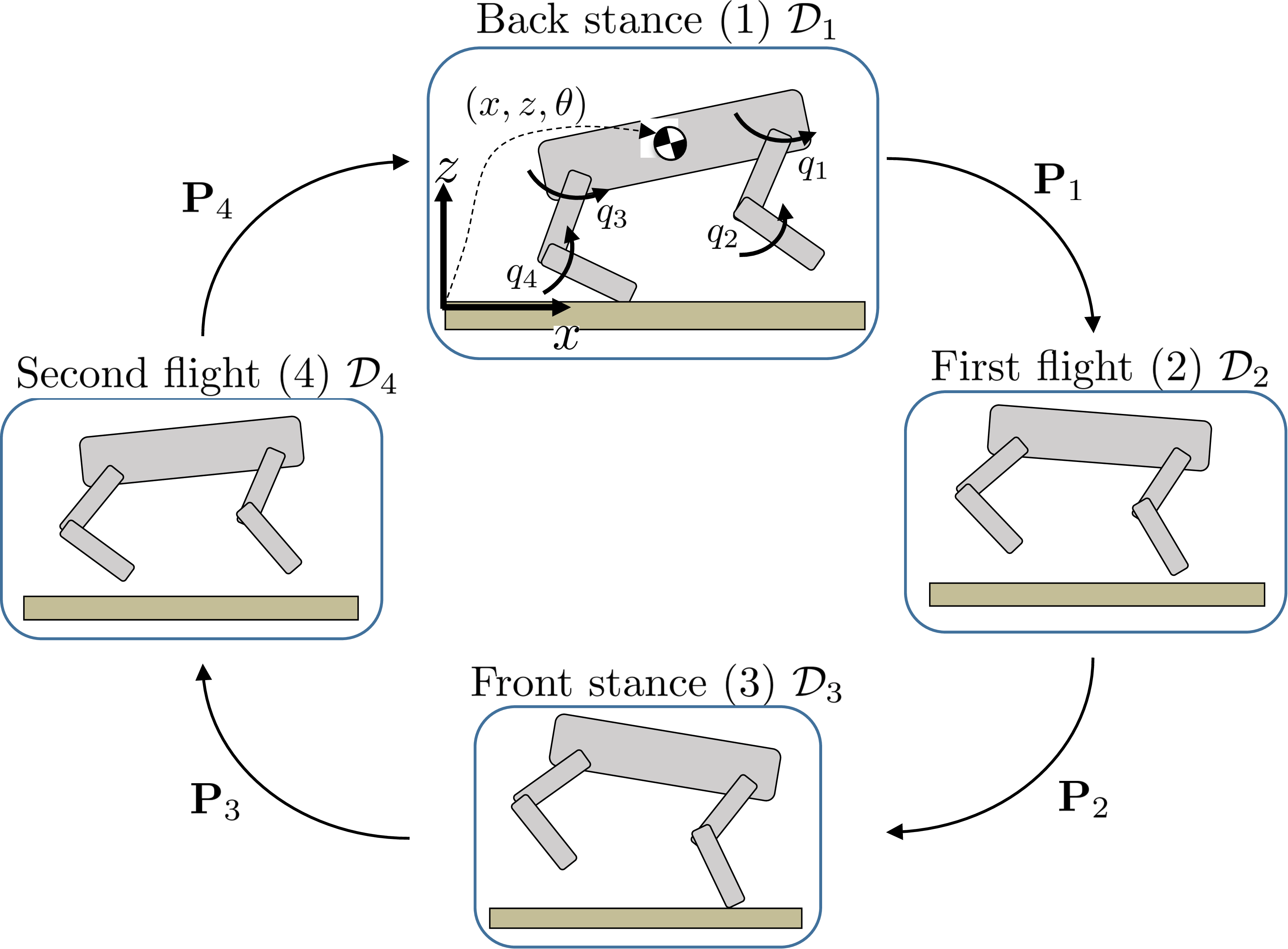}
 \caption{7 DoF planar model~\cite{li2020hybrid} of the MIT Mini-Cheetah~\cite{katz2019mini} showing the 4 modes ($\mathcal{D}_{1} - \mathcal{D}_{4}$) of the optimization problem. $\textbf{P}_{1}$-$\textbf{P}_{4}$ denotes the switching condition for the transition, where $\textbf{P}_{2}$ and $\textbf{P}_{4}$ are the impact conditions.}
 \label{quad}
 \end{figure}
 
\begin{table}[t]
    \centering
    \footnotesize 
    \caption{Elements of the vector used to build the diagonal matrices $ S\in \R^{2 \times 2}$ and $R \in \R^{4 \times 4}$ for different modes}
    \setlength{\tabcolsep}{2pt} 
    \begin{tabularx}{\linewidth}{|p{1.1cm}|*{4}{X|}}
        \hline
        Parameter & Back-Stance & Front-Stance & Flight-1 & Flight-2 \\
        \hline
        \textit{S} & [1,1] & [1,1] & [0,0] & [0,0] \\
        \hline
        \textit{R} & [5.0,0.5,0.1,0.1] & [0.1,0.1,5.0,0.5]&[1.0,1.0,1.0,1.0]& [1.0,1.0,1.0,1.0] \\
        \hline
    \end{tabularx}
    \label{table_weights}
\end{table}

\begin{figure*}[t]
  \includegraphics[width=\textwidth]{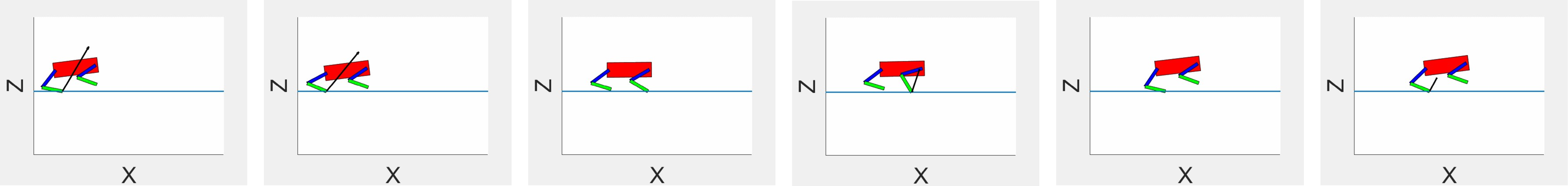}
  \caption{Snapshots of a gait cycle for the 7-DoF quadruped model. The configurations in order are Back-stance (initial), Back-stance (end), First-flight (end), Front-stance (end), Second-flight (end), Back-stance (end). The black arrow shows the contact force ($\lambda$). 
  }
   \label{snapshot_quad}
   \vspace{-7px}
\end{figure*}

An optimization problem for the 7-DoF planar model of the MIT Mini-Cheetah~\cite{katz2019mini} is set up and solved using MDDP. Figure~\ref{quad} shows the four different modes of the problem, which together formulate a single gait cycle. The modes are differentiated by the underlying dynamics (stance or floating base), cost function, and terminal and path constraints. As shown in Fig.~\ref{quad}, the generalized coordiantes are $\q = [x,z,\theta,q_{1}.q_{2},q_{3},q_{4}]$, while the control variables are $\taubar = [\tau_{1},\tau_{2},\tau_{3},\tau_{4}]$. A five-mode problem is set up, beginning from Back-stance to First-Flight, Front-stance, Second-Flight, and finishing with Back-stance. To cast this problem for MDDP, these modes are also the five separate phases, where each phase has its own cost, dynamics, and constraints function. The following sections discuss the details of the optimization problem in different modes (or phases). 

For all the modes, the running cost is defined as:
\begin{equation}
L = (\XMD-\XMD_{ref})\T Q (\XMD-\XMD_{ref}) + \taubar \T R \taubar + \lambda \T S \lambda 
\label{eqn_running_cost}
\end{equation}
where $\XMD = [\q,\qd]$, $\XMD_{ref}$ is the pre-defined reference configuration for each mode determined by heuristics~\cite{li2020hybrid}. The second and the last term minimize quadratic forms of the control effort and the contact forces, respectively, where \textit{R} and \textit{S} are weighting matrices for relative tuning (Tab.~\ref{table_weights}). 
For the flight modes, ($\mathcal{D}_{2},\mathcal{D}_{4}$ in Fig.~\ref{quad}), the terminal cost is also present, and defined as:
\begin{equation}
   L_{f} = (\XMD-\XMD_{ref})\T Q_{f} (\XMD-\XMD_{ref}) 
\end{equation}
The weights on the cost functions defined above are based on rigorous tuning and heuristics for the problem. The dynamics function for the stance modes (Fig.~\ref{quad}) is given by the KKT FD~\eqref{KKT_eqn}, while the Flight modes use Forward Dynamics~\eqref{fwd_dyn2} via the ABA algorithm.
\subsection{Constraints}
MDDP allows to define separate terminal and path constraints for each phase (or mode). For the intermediate back and front stance mode (Fig.~\ref{quad}), there are no terminal constraints, but only path constraints. For the last (fifth mode) Back-Stance, terminal constraints are put on the height $z$ and the pitch $\theta$ variables, otherwise the configuration ends with an unrepeatable geometry. Friction cone path constraints are considered for the stance phases.

The flight phases, on the other hand, employ multiple terminal constraints to tackle the impact conditions ($\textbf{P}_{2}$ and $\textbf{P}_{4}$) in Fig.~\ref{quad}. As an example, for the First-Flight (mode 2), the set of 10 terminal constraints are:
\begin{equation}
    \begin{aligned}
        \psi_{1:7} &= \qd_{s}-impact(\q,\qd_{-}) \\
        \psi_{8} &= foot_z(\q)-0.0 \\
        \psi_{9} &= q_{3}-q_{3,ref} \\
        \psi_{10} &= q_{4}-q_{4,ref} 
    \end{aligned}
    \label{eqn_term_cons}
\end{equation}

Here, $ \psi_{1:7}$ denotes the vector of constraints from the impact event. The quantity $impact(\q,\qd_{-})$ computes the $\qd_{+}$ from the impact dynamics \eqref{im_eqn_1}, and $\qd_{s}$ are the joint velocities for the subsequent front-stance phase. The constraint $\psi_{8}$ in \eqref{eqn_term_cons} forces the $z$ coordinate of the front foot to the ground at impact, while $\psi_{9}$ and $\psi_{10}$ target the non-impact back knee and hip to avoid hindrance with the ground. Along with these, the continuous configuration $\q$ across the impact is forced using the linkage constraints, inherent to MDDP. The FO and SO analytical derivatives for each of the constraints defined in \eqref{eqn_term_cons} are used. For the derivatives of contact forces in friction-cone and positive-contact path constraints, the expressions from Sec.~\ref{sec:KKT_FO} and \ref{sec:KKT_SO} are used. The derivatives of the impact constraint in \eqref{eqn_term_cons} are based on derivatives of impact dynamics (see.~\cite[Sec.~XII]{arxivss_SO}). The transitions $\textbf{P}_{1}$ and $\textbf{P}_{3}$ (in Fig.~\ref{quad}), immediately after the stance modes are modeled using the linkage constraints in MDDP to force continuity in both $\q$ and $\qd$.

\subsection{Problem Set-up \& Solution}
\begin{figure}[t]
 \centering
 \includegraphics[width=0.85 \columnwidth]{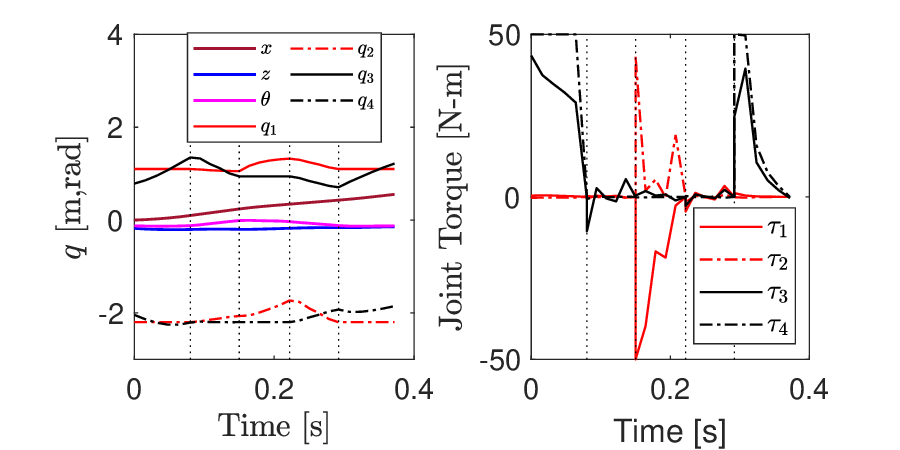}
 \caption{Joint configuration ($\q$), and torques ($\taubar$) for the 5-phase optimization problem. Dashed lines show the states/control for the front and the back knee.}
 \label{states}
 \end{figure}

\begin{figure}[t]
 \centering
 \includegraphics[width=0.8 \columnwidth]{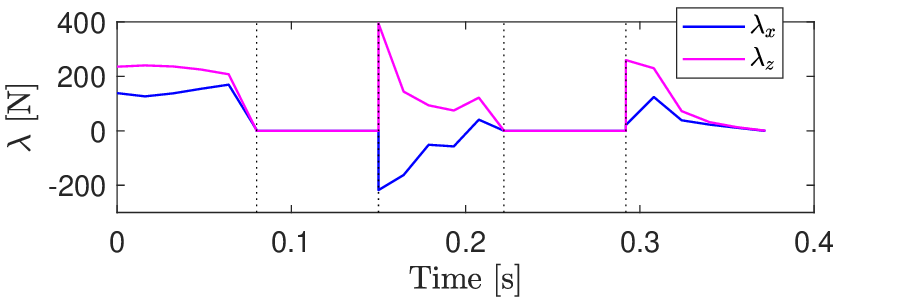}
 \caption{Contact force ($\lambda$) in x and z directions for the 5 phases.}
 \label{contact_force}
 \end{figure}

The 5-phase optimization problem for the quadruped comprises one full gait cycle. Figure~\ref{snapshot_quad} shows snapshots of the initial configuration, followed by the final configuration at the end of each of the 5 phases for the optimal solution. The quadruped starts in the Back-stance and runs forward with a speed of 0.5 m/s\footnote{Animation of the single gait (top) and a double gait (bottom) cycle trajectories can be found at \url{https://youtu.be/C0h6mEpcnAE}.}.  Figure~\ref{states}  shows the resulting optimal state and control trajectory for the problem, with vertical dotted lines to distinguish between each of the 5 phases. The joint torques limits are $-50 \leq \taubar \leq 50$ for all the joints, achieved during the impact events (Fig.~\ref{states}). The initial guess is generated by propagating the initial state for each phase with zero control. The reference state ($\XMD_{ref}$) used in the cost functions is used as the initial state for a phase. This use of $\XMD_{ref}$ helps for convergence since these reference states are also the ``soft'' targets in the running cost function in Eq.~\eqref{eqn_running_cost}. Figure~\ref{contact_force} shows that the contact forces are highest on the impact events, occurring on the phase boundaries of the 3rd and the 5th mode. 

\section{Benefits of Quasi-Newton Approximation}
\label{sec:QN_cases}
As explained in Sec.~\ref{sec:MDDP}, QN approximation is exploited for MDDP to approximate the SO-STM. Different versions of QN, namely QNp are used, with p being chosen by the user. If no QN approximation is used, and the full SO-STM is computed at each iteration, that version is simply named Full-DDP. On the other hand, $\textrm{QN}_{\infty}$ computes the SO-STM only at the first iteration and approximates it after. iLQR depends only on the FO partial derivative information of the dynamics, but is not included in the analysis. 

\begin{figure}[t]
 \centering
 \includegraphics[width=0.80 \columnwidth]{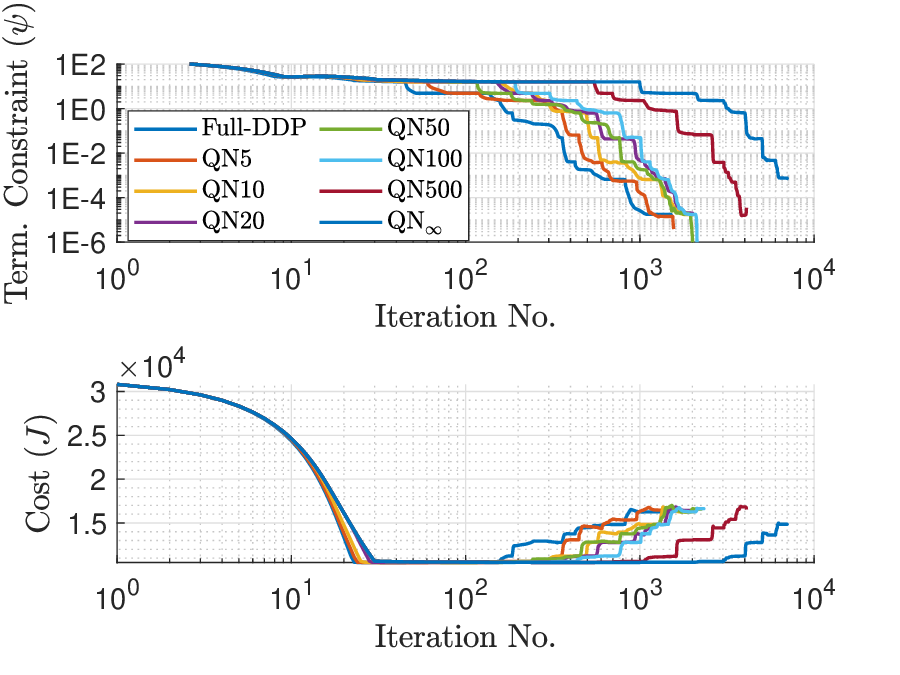}
 \caption{Terminal Constraints and $J$ history for the different versions of QN and Full DDP with number of iterations.}
 \label{term_const}
 \end{figure}

\begin{figure}[t]
 \centering \includegraphics[width=0.8\columnwidth]{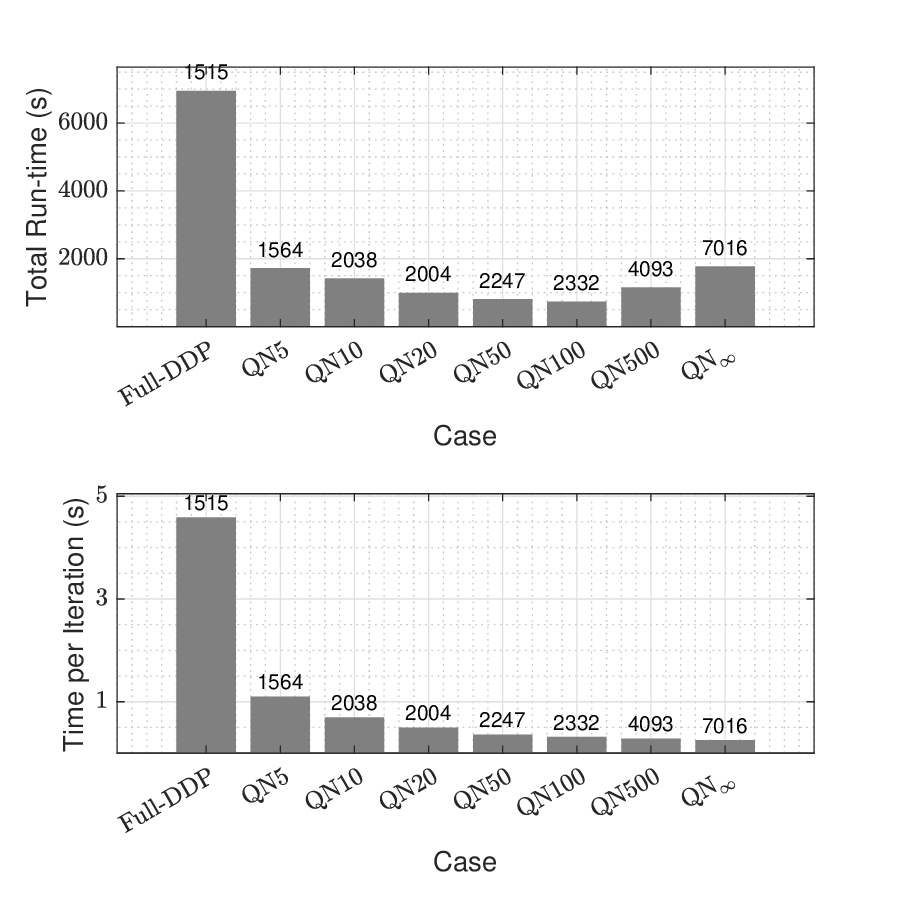}
 \caption{Run-Time and Time per iteration, both in seconds for convergence using different QN cases and full DDP. The iteration count for each case is given on the top of the bar.}
 \label{runtime}
 \end{figure}

Figure~\ref{term_const} shows a comparison of the cost and terminal constraints for different QN cases for the 5-phase quadruped optimization problem. The optimal Cost $J$ in Fig.~\ref{term_const} refers to the sum of the running and the terminal cost. Amongst other criteria~\cite[Sec.~5.3]{pellegrini2020multiple1}, the convergence for MDDP here is measured based on meeting the terminal constraints tolerance to $10^{-3}$, and similarly for the change in $J$ across iterations. From Fig.~\ref{term_const}, the full DDP takes the minimum number of iterations (1515) to converge, followed by QN5. The $\textrm{QN}_{\infty}$ version computes the full SO-STM only at the first iteration and takes the most number of iterations  (7016). Comparing these two extreme cases demonstrate the benefit of including full SO information for DDP. However, as shown in Fig.~\ref{runtime} the total run-time (s), and time per iteration (s) for the full DDP version is, by far, the highest amongst all the cases. $\textrm{QN}_{\infty}$ performs the best on time per iteration, but poorly on the total time due to a high number of iterations. The QN50 and QN100 cases result in the lowest total run-time, and, almost a speedup of 9x and 10x over the full DDP. These order-of-magnitude reductions in runtime are possible because the SO STM calculations account for the bulk of the overall compute effort.
\section{Conclusions}
In this paper, first and second-order partial derivatives of rigid-body dynamics are extended for models with contacts. 
Then, a Multi-Shooting DDP optimizer is used to solve a trajectory optimization for a 7-DoF MIT Mini-Cheetah model for bounded gait motion using. The benefits of using the Quasi-Newton method for approximating the second-order derivatives are presented. As shown with the help of several cases, the speed-ups of using a novel Quasi-Newton over the full DDP method are up to 10x. By using the QN approximation for the vast majority of iterations, the runtimes are reduced to those expected with a FO method. However, by using full SO information every 50 to 100 iterations, the convergence properties are similar to full SO DDP

\section*{Acknowledgement}
The authors thank He Li and John Nganga for helping with the tuning of the optimization problem presented.

\useRomanappendicesfalse

\bibliographystyle{IEEEtran}
\bibliography{main.bib}



\end{document}